# Inference with Possibilistic Evidence

Fengming Song and Ping Liang
College of Engineering
University of California
Riverside, CA 92521-0425

## Abstract

In this paper, the concept of possibilistic evidence which is a possibility distribution as well as a body of evidence is proposed over an infinite universe of discourse. The inference with possibilistic evidence is investigated based on a unified inference framework maintaining both the compatibility of concepts and the consistency of the probability logic.

## 1  INTRODUCTION

In Dempster-Shafer's theory of evidence (Dempster 1967, Shafer 1976) the concept of *basic probability assignment* captures the idea that probabilities should be distributed on subsets of the universe of discourse. A basic probability assignment is obtained instead of a classical probability because of the lack of statistics or lack of full knowledge in expert estimations on all the singletons in the universe of discourse. This lack of knowledge is called *ignorance* (Smets 1991). Ignorance exists universally in the real world.

In practice, it is frequently impossible to carry out statistics directly in order to describe randomness. Even when statistics can be carried out, the number of samples may not be sufficiently large. Estimations from experts are often required or used to estimate uncertainties. Usually, experts cannot provide probability distributions or basic probability assignments based on their expertise to estimate randomness. And the expert estimations of random events are usually expressed in natural language propositions. Following is an example:

**Example**: Suppose that an expert based on his experience and observation estimates that a particular target shooter should be "very accurate". The numbers of the rings actually shot may be 10, 9, 8, ..., it is random. The estimation of the expert can be formulated as a proposition:

$p \triangleq$ "*The ring number shot $(X)$*" is "*very accurate $(\tilde{F})$*";

i.e.  $\quad p \triangleq X \text{ is } \tilde{F}, \; X \in \Theta.$

Here, the expert uses the concept *"very accurate"* to estimate the randomness of the rings actually shot. How the ring numbers 10, 9, 8, ..., are compatible with the concept "very accurate" is fuzzy. Furthermore, the concept "very accurate" must contain some information on the randomness of the ring numbers shot. The question is whether it is possible to extract some information on the probability of the actual ring numbers that will be shot from the fuzzy proposition. We believe that there should exist some connection between the two kinds of uncertainties: the randomness of events and the fuzziness of concepts which are used to estimate the randomness. Although the expert's estimation may have come from statistical experience and observation, but it is not sufficient to specify a basic probability assignment. As a result, he can only use language concepts which have fuzzy membership functions in his mind to describe his estimation of the outcome of a random event. Our goal is to extract as much information as possible from propositions expressed in a natural language.

The above proposition in the example is called a *possibility proposition* in Zadeh's possibility theory (see (Zadeh 1978, 1979) etc.). In the possibility theory, the most important interpretation for the possibility is the compatibility of the value a language variable $(X)$ takes with a fuzzy concept. Zadeh pointed out that there exists the following relation:

$$Probability \leq Possibility$$

However, there should exist tighter and deeper connections between the possibility and the probability when the possibility proposition is used to estimate the outcome of random events. In a possibility proposition, the values taken by the language variable $(X)$ constitute a *possibility distribution* which is a fuzzy subset, i.e. $\tilde{F}$ in the proposition is a fuzzy set. In the theories of random subset coverage or falling shadow proposed by Goodman (Goodman 1982, 1991) and Wang (Wang 1983, 1985), it has been discovered that a fuzzy subset can be represented as the coverage or the falling shadow of a random subset, i.e. the membership function of a fuzzy subset, $\mu_{\tilde{F}}(\theta)$ can be ex-pressed as the coverage or the falling shadow function of a random subset, $\mu_\xi(\theta)$. Combining the theories in (Goodman 1983, 1991, Wang 1983, 1985) with the Dempster-Shafer's evidence theory, it can be shown that, in fact, $\mu_\xi(\theta)$ is the plausibility function on singletons, $Pl_\xi(\theta), \theta \in \Theta$, i.e. $\mu_{\tilde{F}}(\theta) = \mu_\xi(\theta) = Pl_\xi(\theta)$. Therefore, we believe that if the fuzzy subset $\tilde{F}$ in the possibility proposition is used to estimate the outcome of a random event, $\tilde{F}$ must contain implicitly a certain amount of information on a (generalized) probability distribution, i.e., a basic probability assignment.

The information provided by a fuzzy subset $\tilde{F}$ through its membership function is, however, only limited to the plausibility function on singletons. Unfortunately, a basic probability assignment cannot be determined uniquely by the plausibility on singletons. There exist many basic probability assignments whose plausibilities on singletons of the universe of discourse are the same. A rational method is needed to select a basic probability assignment from those having the same plausibility function on singletons to express the information of randomness contained in the fuzzy subset $\tilde{F}$. The selection should be able to reflect the amount of ignorance in a basic probability assignment. In (Song and Liang et.al 1988,1990,1992,1993), we proposed the concept of *possibilistic evidence*. Possibilistic evidence is a possibility distribution (i.e. a fuzzy subset) as well as a body of evidence (i.e. a basic probability assignment).

More important problems occur in inference. Usually, in a decision making process, the information on randomness provided by experts' estimations cannot support the decision conclusion directly and inference is necessary. The inference required is different from fuzzy inference because the probability implication must be considered at all steps, while fuzzy inference only concerns with the compatibility of concepts. Of course, since the information estimating randomness is represented in natural language concepts, the inference should also be consistent with the compatibility of concepts.

This paper investigates the following two problems:

1) Extend the concept of *possibilistic evidence* to an infinite universe of discourse and remove the limitation that the possibility distributions should be normal fuzzy subsets in (Song et al. 1988,1990,1992). The measurability of set-valued mappings will be included in the consideration.

2) Develop a *unified framework* of logic inference with possibilistic evidence. In this framework both the compatibility of concepts and the probabilistic logic should be maintained consistently in inference. This will enable us to gain a better understanding of the connection in inference between the two kinds of uncertainties, randomness and fuzziness. A variety of different kinds of fuzzy operators and probability relations can be investigated systematically to reveal the underlying relation between fuzzy inference and probabilistic inference.

## 2 POSSIBILISTIC EVIDENCE

To extend the evidence theory to an infinite universe of discourse, we define a super-measurable structure on the universe of discourse for the measurability of set-valued mappings.

**Definition 2.1**(The super-measurable structure): Let $\Theta$ be the universe of discourse (frame of discernment), either finite or infinite, and $\mathcal{P}(\Theta)$ be the power space consisting of all the subsets of $\Theta$ (including the empty set $\emptyset$ and $\Theta$ itself). Let $\mathcal{L} \subset \mathcal{P}(\Theta)$ be a subclass of subsets of $\Theta$, which contains $\mathcal{L}_0 = \{\{\theta\}|\theta \in \Theta\}$, the subclass of all the singleton subsets of $\Theta$. Let $\mathcal{B}$ be a $\sigma$-algebra generated by $\mathcal{L}$ over $\Theta$. In addition, for each $B \in \mathcal{L}$, define a $\hat{\mathcal{E}}_B$-type subclass of subsets as $\hat{\mathcal{E}}_B = \{E|E \in \mathcal{B}, E \cap B \neq \emptyset\}$. Let $\tilde{\mathcal{B}}$ be the $\sigma$-algebra generated in $\mathcal{B}$ by the collection of $\hat{\mathcal{E}}_B$'s for all $B \in \mathcal{L}$. Furthermore, the $\sigma$-algebra $\tilde{\mathcal{B}}$ should have the following property: $\hat{\mathcal{E}}_B \in \tilde{\mathcal{B}}$ implies $\hat{\mathcal{E}}_{\bar{B}} \in \tilde{\mathcal{B}}$, where $\bar{B}$ is the complement of $B$ in $\Theta$. Then $(\mathcal{B}, \tilde{\mathcal{B}})$ is called a super-measurable structure over $\Theta$.

Usually, in addition to $\mathcal{L}_0$, $\mathcal{L}$ should include other subsets in $\Theta$. For example, if $\Theta$ is at most countable, it is sufficient to have $\mathcal{L} = \mathcal{L}_0$, and $\mathcal{B}$ will contain the collection of all Borel sets. But if $\Theta$ is not countable, $\mathcal{L}$ should include all the open (or closed) subsets in $\Theta$ so that $\mathcal{B}$ contains the collection of all Borel sets.

**Definition 2.2** A measurable mapping, $\xi : \Omega \to \mathcal{B}$, from a probability field $(\Omega, \mathcal{A}, P)$ to $(\mathcal{B}, \tilde{\mathcal{B}})$, satisfying $\xi^{-1}(\mathcal{C}) = \{\omega|\xi(\omega) \in \mathcal{C}\} \in \mathcal{A}$, for all $\mathcal{C} \in \tilde{\mathcal{B}}$, is called a random subset over the universe of discourse $\Theta$. The set of all the random subsets over $\Theta$ is denoted as $\Xi(\mathcal{A}, \tilde{\mathcal{B}})$.

If a measurable random subset $\xi : \Omega \to \mathcal{B}$ makes $P\{\omega|\xi(\omega) \in \mathcal{B}, \omega \in \Omega\} = 1$ and $K(\xi) = P\{\omega|\xi(\omega) \neq \emptyset, \omega \in \Omega\} > 0$, then $\xi$ is called a basic probability assignment over $\Theta$. All the basic probability assignments constitute a subclass of $\Xi(\mathcal{A}, \tilde{\mathcal{B}})$, denoted as $\Xi_m(\mathcal{A}, \tilde{\mathcal{B}})$.

**Proposition 2.3**: If $\xi : \Omega \to \mathcal{B}$ is a basic probability assignment, then $\{\omega|\xi(\omega) \neq \emptyset, \xi(\omega) \subset A, \omega \in \Omega\}$ and $\{\omega|\xi(\omega) \supset A, \omega \in \Omega\}$ all belong to the $\sigma$-algebra $\mathcal{A}$ for all $A \in \mathcal{B}$. Therefore the following functions are well-defined for each subset $C \subset \Theta$ (ref. (de Fériet 1982)):

$$Bel_\xi(C) = K^{-1}(\xi) \sup_{A \subset C, A \in \mathcal{B}} P\{\omega|\emptyset \neq \xi(\omega) \subset A, \omega \in \Omega\}$$

is called a *belief degree (belief function)*;

$$Pl_\xi(C) = 1 - Bel_\xi(\bar{C})$$

is called a *plausibility degree (plausibility function)*, where $\bar{C}$ is the complement of $C$ in $\Theta$.

$$Q_\xi(C) = \sup_{A \supset C, A \in \mathcal{B}} P\{\omega|\xi(\omega) \supset A, \omega \in \Omega\}$$

is called a *commonality number (commonality function)*.

In a finite universe of discourse and if $K(\xi) = 1$, the above definitions coincide with the original definition in (Shafer 1976). According to Shafer's definitions, $Pl(C)$ and $Bel(C)$ are respectively the *optimistic* and the *conservative estimations* of the probabilities assigned to subset $A$, where $A$ represents a crisp proposition. And $Q(C)$ measures the probabilities that can move freely to every point of $A$ (see (Shafer 1976)).

Next, we present some concepts in possibility theory to prepare the ground for the investigation of the relation between the evidence theory and the possibility theory.

**Definition 2.4**: Let $F_\lambda = \{\theta | \mu_{\tilde{F}}(\theta) \geq \lambda, \theta \in \Theta\}$, $0 < \lambda \leq 1$, be the $\lambda - subset$ of a fuzzy subset $\tilde{F}$. A fuzzy subset $\tilde{F}$ is called as a $\mathcal{B}$-*measurable* one, if all its $\lambda - $ subsets $F_\lambda \in \mathcal{B}$.

We will simply call $\mathcal{B}$-measurable as measurable. Only measurable fuzzy subsets will be considered in this paper.

In Zadeh's *possibility theory* (Zadeh 1978, 1979), *possibility* and *necessity measure* are defined as

$$\Pi_{\tilde{F}}(C) = \text{Poss}\{X \in A\} = \sup_{\theta \in A} \mu_{\tilde{F}}(\theta), \theta \in \Theta;$$

$$N_{\tilde{F}}(C) = 1 - \text{Poss}\{X \in \bar{A}\} = 1 - \sup_{\theta \in \bar{A}} \mu_{\tilde{F}}(\theta), \theta \in \Theta.$$

These definitions are only suitable for *normal fuzzy subsets* (i.e. $F_1 \neq \emptyset$, implying $\sup_{\theta \in \Theta} \mu_{\tilde{F}}(\theta) = 1$). Otherwise, for some $A \subset \Theta$, there may be $\Pi_{\tilde{F}}(C) < N_{\tilde{F}}(C)$, yielding a contradiction. Therefore we redefine them as

**Definition 2.5**

$$\Pi_{\tilde{F}}(C) = \frac{\sup_{\theta \in C} \mu_{\tilde{F}}(\theta)}{\sup_{\theta \in \Theta} \mu_{\tilde{F}}(\theta)}, \; N_{\tilde{F}}(C) = 1 - \frac{\sup_{\theta \in \bar{C}} \mu_{\tilde{F}}(\theta)}{\sup_{\theta \in \Theta} \mu_{\tilde{F}}(\theta)};$$

and

$$Q_{\tilde{F}}(C) = \inf_{\theta \in C} \mu_{\tilde{F}}(\theta)$$

where $Q_{\tilde{F}}(C)$ is the *commonality number (commonality function)* of $\tilde{F}$.

We use the falling shadow or coverage function of random subsets as a tool to explore the connection between the evidence theory and the possibility theory. The formal definition of a falling shadow or coverage function of a random subset is given below.

**Definition 2.6**: Assume $\xi : \Omega \to \mathcal{B}$ is a basic probability assignment over $\Theta$. Define

$$\mu_\xi(\theta) = P\{\omega | \xi(\omega) \ni \theta, \; \omega \in \Omega\}, \theta \in \Theta.$$

$\mu_\xi(\theta)$ is called the falling shadow or coverage function of $\xi$.

From **Definition 2.1** and **2.2**, it is certain that $\mu_\xi(\theta)$ exists for all $\theta \in \Theta$.

It can be proved that, for a fuzzy subset $\tilde{F}$, there exists a falling shadow (or coverage) function of random subsets $\mu_\xi(\theta)$ representing its membership function $\mu_{\tilde{F}}(\theta)$, and this correspondence is many-to-one. We have

**Definition 2.7** $\xi/\sim$ is called an *equivalent subclass* of falling shadow (or coverage) of random subsets for a (measurable) fuzzy subset $\tilde{F}$, if it consists of all the (measurable) random subsets having the following properties: if $\xi \in \xi/\sim$, then

$$\mu_\xi(\theta) = \mu_{\tilde{F}}(\theta), \; \theta \in \Theta, \; \text{and} \; K(\xi) = \sup_{\theta \in \Theta} \mu_{\tilde{F}}(\theta).$$

Under certain mathematical assumption (see **Assumption 2.11**), we can prove that an equivalent subclass of falling shadow (or coverage) is not empty and, in general, it contains more than one random subset. We have

**Theorem 2.8** (The fundamental theorem of possibilistic evidence): There exists a $\xi_0 \in \xi/\sim$, whose images in $\Theta$ are $\lambda$-subsets of $\tilde{F}$, $F'_\lambda s$, moreover, for all $\xi \in \xi/\sim$, there are

$$\Pi_{\tilde{F}}(C) = Pl_{\xi_0}(C) \leq Pl_\xi(C);$$
$$N_{\tilde{F}}(C) = Bel_{\xi_0}(C) \geq Bel_\xi(C)$$

and

$$Q_{\tilde{F}}(C) = Q_{\xi_0}(C) \geq Q_\xi(C)$$

for all $C \subset \Theta$. Therefore $\xi/\sim$ can be denoted as $\xi_0/\sim$.

**Note**: The condition $K(\xi) = \sup_{\theta \in \Theta} \mu_{\tilde{F}}(\theta)$ can be removed for the third inequality.

The basic probability assignment $\xi_0 : \Omega \to \mathcal{B}$ and the corresponding belief, plausibility and commonality functions are altogether called the *possibilistic evidence* induced from the possibility proposition $p \stackrel{\triangle}{=} X$ is $\tilde{F}$, $X \in \Theta$.

From **Theorem 2.8**, we can conclude that the choice of the possibilistic evidence from the equivalent subclass of falling shadow (or coverage) of random subsets induced by the possibility distribution $\tilde{F}$ assures that the maximum amount of information on randomness from the possibility proposition is extracted. This is based on the belief that the language proposition contains valuable information of expert knowledge estimating randomness.

The following definitions are needed to investigate the inference of possibilistic evidence.

**Definition 2.9**: Suppose $(\Omega, \mathcal{A}, P)$ is a probability field, a $\mathcal{A}$-*division* over $\Omega$ is defined as a subclass of subsets of $\Omega$: $d = \{D_i | i \in I\}$, where $I$ is at most countable, satisfying

$$D_i \in \mathcal{A} \; (i \in I), \; \bigcup_{i \in I} D_i = \Omega, \; D_i \cap D_j = \emptyset, \; \text{for} \; i \neq j.$$

The collection of all $\mathcal{A}$-division over $\Omega$ is denoted as $\mathcal{D} = \mathcal{D}(\Omega, \mathcal{A})$. The *normal net* of $(\Omega, \mathcal{A}, P)$ is a division series $\mathcal{D}^* = \{d^{(n)}\} \subset \mathcal{D}(\Omega, \mathcal{A})$:

$$d^{(n)} = \{D_{i_1 \ldots i_n} | i_k = 0, 1, k = 1, \ldots, n\},$$

$$D_{i_1\ldots i_n} \in \mathcal{A}, \ P(D_{i_1\ldots i_n}) = \frac{1}{2^n},$$
$$D_{i_1\ldots i_{n-1}0} \cup D_{i_1\ldots i_{n-1}1} = D_{i_1\ldots i_{n-1}}; n = 1, 2, \ldots$$

**Definition 2.10**: $\mathcal{A}$ is said to be sufficient to $\mathcal{B}$, if $Proj_\Omega(\mathcal{A} \times \mathcal{B}) \triangleq \{Proj_\Omega D | D \in \mathcal{A} \times \mathcal{B}\} = \mathcal{A}$, where $Proj_\Omega D \triangleq \{\omega | \exists \theta \in \Theta : (\omega, \theta) \in D\}$ is the projection of $D$ into $\Omega$.

We assume that all the probability fields in this paper satisfy the following assumption:

**Assumption 2.11** The probability field $(\Omega, \mathcal{A}, P)$ has its *fixed* normal net and $\mathcal{A}$ is sufficient to $\mathcal{B}$.

## 3  INFERENCE WITH POSSIBILISTIC EVIDENCE

### 3.1  NEGATION, SYNONYMOUS AND ANTONYMOUS CLASSES OF CONCEPTS

**Definition 3.1.1**: $\xi : \Omega \to \mathcal{B}$ is a set-valued mapping, and
$$G_\xi \triangleq \{(\omega, \theta) | \xi(\omega) \ni \theta\}$$
is called the graph of $\xi$.

**Lemma 3.1.2**: Let
$\xi(\omega) = G|_{\cdot \omega}$, where $G|_{\cdot \omega} \triangleq \{\theta|(\omega, \theta) \in G\}$, and $G \in \mathcal{A} \times \mathcal{B}$, then $\xi : \Omega \to \mathcal{B}$ is a (measurable) random subset.

**Definition 3.1.3**: $G \in \mathcal{A} \times \mathcal{B}, G|_{\cdot \theta} = \{\omega|(\omega, \theta) \in G\}$ is called the cutting shadow of $G$ at $\theta$.

If $\tilde{F}$ is a measurable fuzzy subset, then its membership function $\mu_{\tilde{F}}$ is a $(\mathcal{B}\text{-})$ measurable real function. Therefore, there exists a sequence of $(\mathcal{B}\text{-})$ simple functions $f_n(\theta) \uparrow \mu_{\tilde{F}}(\theta)$: $f_n(\theta) \triangleq \sum_{k=1}^{2^n} \frac{k}{2^n} \mathcal{X}_{\tilde{F}, \frac{k}{2^n}}(\theta)$ where
$$\mathcal{X}_{\tilde{F},\frac{k}{2^n}}(\theta) = \begin{cases} 1, & \text{if } \mu_{\tilde{F}}(\theta) \in [\frac{k}{2^n}, \frac{k+1}{2^n}]; \\ 0, & \text{otherwise.} \end{cases}$$

The above sequence of simple functions $f_n(\theta)$ are defined on the normal net $\mathcal{D}^*$ of $(\Omega, \mathcal{A}, P)$ for $n \geq 1$. The following transformation maps the binary order of the normal net to a decimal order.
$$\zeta : (i_1, \ldots, i_n) \longmapsto k \triangleq (i_1 \cdot 2^{n-1} + i_2 \cdot 2^{n-2} + \cdots + i_n) + 1,$$
where $\zeta$ is *one to one*. Denote
$$B_k^{(n)} \triangleq \bigcup_{i=1}^k D_{\rho(i)} \quad (k = 1, 2, \ldots, 2^n), \text{ where } \rho = \zeta^{-1}.$$

Then $\quad B_k^{(n)} \in \mathcal{A}, \quad P(B_k^{(n)}) = \frac{k}{2^n}.$

The sequence of simple functions $f_n(\theta)$ can be constructed using the following graph:
$$G_{\tilde{F}}^{(n)} \triangleq \bigcup_{k=1}^{2^n} \{B_k^{(n)} \times \{\theta | \mu_{\tilde{F}}(\theta) \geq \frac{k}{2^n}\}\}.$$

It is obviously that $f_n(\theta) = P(G_{\tilde{F}}^{(n)}|_{\cdot \theta})$. If $n_1 < n_2$, then $G_{\tilde{F}}^{(n_1)} \subset G_{\tilde{F}}^{(n_2)}$. Let $G_{\tilde{F}}^* = \lim_{n \to \infty} G_{\tilde{F}}^{(n)} = \bigcup_{n=1}^\infty G_{\tilde{F}}^{(n)}$. This assures that $G_{\tilde{F}}^* \in \mathcal{A} \times \mathcal{B}$ and $\mu_{\tilde{F}}(\theta) = P(G_{\tilde{F}}^*|_{\cdot \theta})$. The possibilistic evidence in **Theorem 2.8** can be obtained using the $G_{\tilde{F}}^*$:

**Proposition 3.1.4**: Let $\xi_0(\omega) = G_{\tilde{F}}^*|_{\cdot \omega}$, then $\xi_0 : \Omega \to \mathcal{B}$ is the possibilistic evidence of **Theorem 2.8**.

It is obvious that $G_{\tilde{F}}^*$ is the graph of $\xi_0$, i.e., $G_{\tilde{F}}^* = \{(\omega, \theta)|\xi_0(\omega) \ni \theta\}$.

Let $\bar{G}_\xi = (\Omega \times \Theta)/G_\xi$ and $\bar{\xi}(\omega) = \bar{G}_\xi|_{\cdot \omega}$, then $\bar{\xi}$ is called the complement of $\xi$. If $\xi \in \xi/\sim$ in **Definition 2.7**, We have

**Proposition 3.1.5**: $\bar{\xi} : \Omega \to \mathcal{B}$ is a (measurable) random subset with its falling shadow (or coverage) $\mu_{\bar{\xi}}(\theta) = 1 - \mu_{\tilde{F}}(\theta) = \mu_{\bar{\tilde{F}}}(\theta), \theta \in \Theta$. Then $\bar{\xi}$ represents the fuzzy subset $\bar{\tilde{F}}$, the complement of the fuzzy subset $\tilde{F}$, i.e. it represents the *negation proposition* $\neg p \triangleq X \text{ is } (not\ \tilde{F}), X \in \Theta$. The operation $\xi \to \bar{\xi}$ is called *negation*.

It is obvious that $\bar{\bar{\xi}} = \xi$. Morever, there is

**Theorem 3.1.6**: If $\xi_0$ is the possibilistic evidence in the equivalent subclass induced from $\tilde{F}$, $\bar{\xi}/\sim$ is the equivalent subclass of falling shadow (or coverage) induced from $\bar{\tilde{F}}$, i.e., for all $\bar{\xi} \in \bar{\xi}/\sim$, there is $\mu_{\bar{\xi}}(\theta) = \mu_{\bar{\tilde{F}}}(\theta) = 1 - \mu_{\tilde{F}}(\theta), \theta \in \Theta$, and $K(\bar{\xi}) = \sup_{\theta \in \Theta} \mu_{\bar{\tilde{F}}}(\theta) = 1 - \inf_{\theta \in \Theta} \mu_{\tilde{F}}(\theta)$. Then $\bar{\xi}_0$, induced from $\xi_0$ by negation, is the possibilistic evidence in $\bar{\xi}/\sim$, i.e., for all $\bar{\xi} \in \bar{\xi}/\sim$, there are
$$\Pi_{\bar{\tilde{F}}}(C) = Pl_{\bar{\xi}_0}(C) \leq Pl_{\bar{\xi}}(C);$$
$$N_{\bar{\tilde{F}}}(C) = Bel_{\bar{\xi}_0}(C) \geq Bel_{\bar{\xi}}(C)$$
for all $C \subset \Theta$. Therefore $\bar{\xi}/\sim$ can be denoted as $\bar{\xi}_0/\sim$.

In addition, there is

**Proposition 3.1.7**: For all $\omega \in \Omega$, there is $\bar{\xi}(\omega) = \overline{\xi(\omega)}$, where $\overline{\xi(\omega)} = \Theta/\xi(\omega)$.

**Theorem 3.1.8**: If $\xi_0$ is the possibilistic evidence in the equivalent subclass induced from $\tilde{F}$, then for all $\xi$ having the property $\mu_\xi(\theta) = \mu_{\tilde{F}}(\theta), \theta \in \Theta$, there is
$$Q_{\bar{\tilde{F}}}(C) = Q_{\bar{\xi}_0}(C) \geq Q_{\bar{\xi}}(C)$$
for all $C \subset \Theta$, where $\bar{\xi}$ is obtained from $\xi$ through negating. Furthemore, we have the relation
$$Q_\xi(C) + K(\bar{\xi})Pl_{\bar{\xi}}(C) = 1.$$

In some universe of discourse $\Theta$, there exist two classes of concepts whose implications are antonymous to each other, e.g., $\{small,\ quite\ small,\ very\ small,\ extremely\ small,\ \ldots\}$ versus $\{large,\ quite\ large,\ very\ large,\ extremely\ large,\ \ldots\}$. Given two concepts represented by two fuzzy subsets $\tilde{F}_1$ and $\tilde{F}_2$, if both $\tilde{F}_1$ and $\tilde{F}_2$ belong to the same class of concepts, the two concepts are said to be *synonymous* in implication, otherwise, they are said to be *antonymous* in implication. We first select one of the two classes, e.g., the class containing $\tilde{F}_1$. We take a body of possibilistic evidence $\xi_1$ as the representation of $\tilde{F}_1$, defining a set-valued mapping from $(\Omega_1, \mathcal{A}_1, P_1)$ to $(\mathcal{B}, \tilde{\mathcal{B}})$: $\xi_1(\omega_1) \triangleq G^*_{\tilde{F}_1}|_{\cdot \omega_1} : \Omega_1 \to \mathcal{B}$ as in **Proposition 3.1.4**. If $\tilde{F}_2$ is synonymous to $\tilde{F}_1$, define a set-valued mapping from $(\Omega_2, \mathcal{A}_2, P_2)$ to $(\mathcal{B}, \tilde{\mathcal{B}})$, $\xi_2 : \Omega_2 \to \mathcal{B}$ in the same way as $\xi_1$, $\xi_2(\omega_2) \triangleq G^*_{\tilde{F}_2}|_{\cdot \omega_2}$. If $\tilde{F}_2$ is antonymous to $\tilde{F}_1$, then we firstly take $\bar{\xi}_2(\omega_2) \triangleq G^*_{\tilde{F}_2}|_{\cdot \omega_2}$ as in **Proposition 3.1.4**, after negating $\bar{\xi}_2 \to \bar{\bar{\xi}}_2$ as in **Proposition 3.1.5**, $\xi_2(\omega_2) = \bar{\bar{\xi}}_2(\omega_2) = \bar{G}^*_{\tilde{F}_2}|_{\cdot \omega_2}$ is obtained, where $\bar{G}^*_{\tilde{F}_2} = (\Omega_2 \times \Theta)/G^*_{\tilde{F}_2}$. From **Theorem 3.1.6** and **3.1.8**, $\xi_2 = \bar{\bar{\xi}}_2$ is also a body of possibilistic evidence.

After making this classification, a series of interesting inference results can be obtained. In this paper, we only investigate the inference involving concepts which belong to either synonymous or antonymous classes as extreme cases. We believe that an appropriate selection of the universe of discourse can ensure the concepts involved in inference to be limited to beloging to two classes which are antonymous in implication to each other. The rest of the concepts involved can be obtained through combination. We will investigate this topic in another papers.

### 3.2 UNION AND INTERSECTION OPERATIONS, CORRELATION OF THE KNOWLEDGE INFORMATION SOURCES

When two possibility propositions $p_1 \triangleq X$ is $\tilde{F}_1, X \in \Theta$ and $p_2 \triangleq X$ is $\tilde{F}_2, X \in \Theta$ are used to estimate the randomness of the values taken by the same language variable $X$, the results of *union* and *intersection operations* are the estimation of randomness of the values the variable $X$ takes induced by the union and intersection possibility propositions $\check{p}_{12} \triangleq X$ is $(\tilde{F}_1$ or $\tilde{F}_2), X \in \Theta$ and $\hat{p}_{12} \triangleq X$ is $(\tilde{F}_1$ and $\tilde{F}_2), X \in \Theta$, respectively. The two bodies of possibilistic evidence $\xi_1$ and $\xi_2$ representing $\tilde{F}_1$ and $\tilde{F}_2$, respectively, should be considered as being defined on their own individual probability fields $(\Omega_1, \mathcal{A}_1, P_1)$ and $(\Omega_2, \mathcal{A}_2, P_2)$ respectively. The two probability fields can be considered as two knowledge information sources providing $p_1$ and $p_2$. Of course, there is a special case that $(\Omega_1, \mathcal{A}_1, P_1)$ and $(\Omega_2, \mathcal{A}_2, P_2)$ are the same one. Therefore, the results of union and intersection operations, $\check{\xi}_{12} \triangleq \xi_1 \vee \xi_2$ and $\hat{\xi}_{12} \triangleq \xi_1 \wedge \xi_2$, should be defined on the combined probability field $(\Omega_1 \times \Omega_2, \mathcal{A}_1 \times \mathcal{A}_2, P_{12})$, where $\Omega_1 \times \Omega_2$ is the product sample space, $\mathcal{A}_1 \times \mathcal{A}_2$ is the product $\sigma$-algebra and $P_{12}$ is the combined probability measure. $\check{\xi}_{12}$ and $\hat{\xi}_{12}$: $\Omega_1 \times \Omega_2 \to \mathcal{B}$ are obtained as follows:

**Definition 3.2.1**: $\xi_1 : \Omega_1 \to \mathcal{B}$ and $\xi_2 : \Omega_2 \to \mathcal{B}$ are two set-valued mappings. Define

$$G_{\xi_1 \vee \xi_2} \triangleq \{(\omega_1, \omega_2, \theta) | \xi_1(\omega_1) \cup \xi_2(\omega_2) \ni \theta\}$$

and    $G_{\xi_1 \wedge \xi_2} \triangleq \{(\omega_1, \omega_2, \theta) | \xi_1(\omega_1) \cap \xi_2(\omega_2) \ni \theta\}.$

as *graphs* of $\check{\xi}_{12} \triangleq \xi_1 \vee \xi_2$ and $\hat{\xi}_{12} \triangleq \xi_1 \wedge \xi_2$, respectively. Then

$$\check{\xi}_{12}(\omega_1, \omega_2) = G_{\xi_1 \vee \xi_2}|_{\cdot(\omega_1, \omega_2)} : \Omega_1 \times \Omega_2 \to \mathcal{B}$$
$$\hat{\xi}_{12}(\omega_1, \omega_2) = G_{\xi_1 \wedge \xi_2}|_{\cdot(\omega_1, \omega_2)} : \Omega_1 \times \Omega_2 \to \mathcal{B}$$

are the *union* and *intersection* of $\xi_1$ and $\xi_2$, respectively. The operations $(\xi_1, \xi_2) \to \check{\xi}_{12}$ and $(\xi_1, \xi_2) \to \hat{\xi}_{12}$ are called *union* and *intersection* respectively.

The structure of the combined probability field affects the results of inference. Especially, different combined probability measure $P_{12}$'s produce different results of inference. First, we introduce the following definitions:

**Definition 3.2.2**: Let $\mathcal{D}_1^* = \{d_1^{(n)}\} \subset \mathcal{D}(\Omega_1, \mathcal{A}_1)$ and $\mathcal{D}_2^* = \{d_2^{(n)}\} \subset \mathcal{D}(\Omega_2, \mathcal{A}_2)$ be the *normal nets* of $(\Omega_1, \mathcal{A}_1, P_1)$ and $(\Omega_2, \mathcal{A}_2, P_2)$, respectively (see **Definition 2.10**), where $d_1^{(n)} = \{D^{(1)}_{i_1 \ldots i_n} | i_k = 0, 1; k = 1, \ldots n\}$ and $d_2^{(n)} = \{D^{(2)}_{j_1 \ldots j_n} | j_r = 0, 1; r = 1, \ldots n\}$. A *combined net* of $(\Omega_1 \times \Omega_2, \mathcal{A}_1 \times \mathcal{A}_2, P_{12})$ is a division series $\mathcal{D}_{12}^* = \{d_{12}^{(n)}\} \subset \mathcal{D}(\Omega_1 \times \Omega_2, \mathcal{A}_1 \times \mathcal{A}_2)$ as follows:

$$d_{12}^{(n)} = \{(D^{(1)}_{i_1 \ldots i_n}, D^{(2)}_{j_1 \ldots j_n}) | i_k, j_r = 0, 1; k, r = 1, \ldots n\},$$

$$D^{(1)}_{i_1 \ldots i_n} \in \mathcal{A}_1, D^{(2)}_{j_1 \ldots j_n} \in \mathcal{A}_2,$$

$$P_{12}\{(D^{(1)}_{i_1 \ldots i_n}, \Omega_2)\} = P_1(D^{(1)}_{i_1 \ldots i_n}) = \frac{1}{2^n},$$

$$P_{12}\{(\Omega_1, D^{(2)}_{j_1 \ldots j_n})\} = P_2(D^{(2)}_{j_1 \ldots j_n}) = \frac{1}{2^n},$$

$$(D^{(1)}_{i_1 \ldots i_{n-1} 0}, D^{(2)}_{j_1 \ldots j_{n-1} 0}) \cup (D^{(1)}_{i_1 \ldots i_{n-1} 0}, D^{(2)}_{j_1 \ldots j_{n-1} 1}) \cup$$
$$D^{(1)}_{i_1 \ldots i_{n-1} 1}, D^{(2)}_{j_1 \ldots j_{n-1} 0}) \cup (D^{(1)}_{i_1 \ldots i_{n-1} 1}, D^{(2)}_{j_1 \ldots j_{n-1} 1})$$
$$= (D^{(1)}_{i_1 \ldots i_{n-1}}, D^{(2)}_{j_1 \ldots j_{n-1}}), \quad n = 1, 2, \ldots$$

It is apparent that for any $n$, there should be

$$\bigcup_{i_1 \ldots i_n, j_1 \ldots j_n} (D^{(1)}_{i_1 \ldots i_n}, D^{(2)}_{j_1 \ldots j_n}) = \Omega_1 \times \Omega_2,$$

$$(D^{(1)}_{i_1 \ldots i_n}, D^{(2)}_{j_1 \ldots j_n}) \cap (D^{(1)}_{k_1 \ldots k_n}, D^{(2)}_{r_1 \ldots r_n}) = \emptyset,$$
if $i_1 \ldots i_n \neq k_1 \ldots k_n$ or $j_1 \ldots j_n \neq r_1 \ldots r_n$.

The combined probability field $(\Omega_1 \times \Omega_2, \mathcal{A}_1 \times \mathcal{A}_2, P_{12})$ should also satisfy **Assumption 2.11**, i.e., it has its combined net $\mathcal{D}_{12}^*$, and $\mathcal{A}_1 \times \mathcal{A}_2$ is sufficient to $\mathcal{B}$.

There are three specific combined probability measures $P_{12}$ that are of particular interest:

**Definition 3.2.3**: If for any $n \geq 1$ and every element of $d_{12}^{(n)} \in \mathcal{D}_{12}^* = \{d_{12}^{(n)}\}$,

1) if $P_{12}\{(D_{i_1...i_n}^{(1)}, D_{j_1...j_n}^{(2)})\} = \begin{cases} \frac{1}{2^n}, & \text{if } i_1...i_n = j_1...j_n, \\ 0, & \text{otherwise}; \end{cases}$

$P_{12}$ is said to be *positively correspondent* and is denoted as $P_{12}^+$.

2) if $P_{12}\{(D_{i_1...i_n}^{(1)}, D_{j_1...j_n}^{(2)})\} = P_1(D_{i_1...i_n}^{(1)})P_2(D_{j_1...j_n}^{(2)})$

$$= \frac{1}{4^n} \quad \text{for all } i_1...i_n, j_1...j_n;$$

$P_{12}$ is said to be *independent* and is denoted as $P_{12}^I$.

3) if $P_{12}\{(D_{i_1...i_n}^{(1)}, D_{j_1...j_n}^{(2)})\} = \begin{cases} \frac{1}{2^n}, & \text{if } j_k = 1 - i_k, \\ & k = 1,...n, \\ 0, & \text{otherwise}; \end{cases}$

$P_{12}$ is said to be *negatively correspondent* and is denoted as $P_{12}^-$.

All the results of inference induced from $P_{12}^+$, $P_{12}^I$ and $P_{12}^-$ will be denoted by $+$, $I$ and $-$, e.g., $\mu_{\xi_1 \wedge \xi_2}^+(\theta)$, $K^I(\xi_1 \vee \xi_2)$, $Bel_{\xi_1 \to \xi_2}^-(C)$, etc. All results without superscript represent those induced from other $P_{12}'s$.

The normal net of a probability field is a sequence of finer and finer subdivisions of this probability space. It induces an order on the sample space. The order is fixed according to **Assumption 2.11**. This order induces an embedding of the sample space into the unit interval. Having two probability spaces of this kind, their normal nets define an embedding of the product sample space into the unit square. The measures $P_{12}^+$, $P_{12}^-$ and $P_{12}^I$ correspond to the uniform distribution on the main diagonal, the uniform distribution on the secondary diagonal, and the uniform distribution over the unit square, respectively. If we consider that the probability fields are the sources providing the knowledge information, the way the combined probability is distributed reflects the correlation between the two information sources. If the information sources are individual experts, the correlation of the information sources then depends on the background, experience, personality, etc. of the experts. Considering the fuzzy subsets included in the possibility propositions as "*soft constraints*" on the values taken by the language variable ($X$), the order imposed by the normal net on a probability field represents the degree of "*tightness*" or "*looseness*" of the soft constraints. For two concepts which are synonymous in implication, $P_{12}^+$ implies that the opinions of the two experts on the degree of the soft constraints are similar, and $P_{12}^-$ means that the opinions are adversary, and vise versa for concepts which are antonymous in implication. Therefore, the correlation of the knowledge information sources represented by the relation between the orders of constraints is the correlation of the experts' perception of the concepts. Based on this understanding, we have

**Proposition 3.2.4**: If $\xi_1$ and $\xi_2$ are induced from two concepts $\tilde{F}_1$ and $\tilde{F}_2$ which are synonymous in implication, there is
$$K^-(\check{\xi}_{12}) \geq K(\check{\xi}_{12}) \geq K^+(\check{\xi}_{12});$$
and if $\xi_1$ and $\xi_2$ are antonymous in implication, the direction of the above inequalities should be reversed.

**Theorem 3.2.5**: If $K^-(\check{\xi}_{12}) = K^+(\check{\xi}_{12})$ (then all $K(\check{\xi}_{12}) = constant$.) and $\xi_1$ and $\xi_2$ are induced from two concepts $\tilde{F}_1$ and $\tilde{F}_2$ which are synonymous in implication, there is
$$Pl_{\check{\xi}_{12}}^-(C) \geq Pl_{\check{\xi}_{12}}(C) \geq Pl_{\check{\xi}_{12}}^+(C),$$
$$Bel_{\check{\xi}_{12}}^-(C) \leq Bel_{\check{\xi}_{12}}(C) \leq Bel_{\check{\xi}_{12}}^+(C);$$
for all $C \subset \Theta$. If $\xi_1$ and $\xi_2$ are induced from two concepts $\tilde{F}_1$ and $\tilde{F}_2$ which are antonymous in implication, the direction of the inequalities should be reversed.

The logical implication of **Theorem 3.2.5** is interesting and agrees with intuition. A special case of $P_{12}^+$ is that $(\Omega_1, \mathcal{A}_1, P_1) = (\Omega_2, \mathcal{A}_2, P_2)$ which implies that the two knowledge information sources (the two experts) are the same one. In this case, the implication of **Theorem 3.25** is obvious.

The condition required in **Theorem 3.2.5** is satisfiable, e.g., if either $\tilde{F}_1$ or $\tilde{F}_2$ is a normal fuzzy subset, there will be $K^-(\check{\xi}_{12}) = K^+(\check{\xi}_{12}) = 1$.

**Corollary 3.2.6**: For concepts $\tilde{F}_1$ and $\tilde{F}_1$ which are synonymous in implication, there is
$$\mu_{\check{\xi}_{12}}^-(\theta) \geq \mu_{\check{\xi}_{12}}(\theta) \geq \mu_{\check{\xi}_{12}}^+(\theta)$$
for all $\theta \in \Theta$. And if $\tilde{F}_1$ and $\tilde{F}_2$ are antonymous in implication, the direction of the inequalities should be reversed.

For the intersection operation $\hat{\xi}_{12} \triangleq \xi_1 \wedge \xi_2$, unfortunately, there exists no similar relation like those in **Proposition 3.2.4** and **Theorem 3.2.5**. However, we have:

**Theorem 3.2.7**: If $\xi_1$ and $\xi_2$ are induced from two concepts $\tilde{F}_1$ and $\tilde{F}_2$ which are synonymous in implication, there is
$$Q_{\hat{\xi}_{12}}^-(C) \leq Q_{\hat{\xi}_{12}}(C) \leq Q_{\hat{\xi}_{12}}^+(C);$$
and for the two concepts which are antonymous to each other, the direction of the inequalities is reversed.

**Corollary 3.2.8**: For two concepts $\tilde{F}_1$ and $\tilde{F}_2$ which are synonymous in implication, there is
$$\mu_{\hat{\xi}_{12}}^-(\theta) \leq \mu_{\hat{\xi}_{12}}(\theta) \leq \mu_{\hat{\xi}_{12}}^+(\theta),$$
for all $\theta \in \Theta$. And if $\tilde{F}_1$ and $\tilde{F}_1$ are antonymous in implication, the direction of the inequalities should be reversed.

The term in the center of all the above inequalities in the propositions and theorems includes the case induced by $P_{12}^I$ as a special one. The case for more than two bodies of evidence will be discussed in another paper.

## 3.3 CONDITIONING, EXTENDING OF THE UNIVERSE OF DISCOURSE

A conditional inference proposition $\vec{p}_{12} \triangleq$ If $X$ is $\tilde{F}_1$, then $Y$ is $\tilde{F}_2$, $X \in \Theta_1$, $Y \in \Theta_2$ describes the logical relation between the two language variables $X$ and $Y$. The two variables take their values in their own universes of discourse $\Theta_1$ and $\Theta_2$ ($\Theta_1$ and $\Theta_2$ may be the same one). If $\Theta_1$ and $\Theta_2$ are different, the investigation should be made in the Cartesian product of the two universes of discourse, $\Theta_1 \times \Theta_2$, and a super-measurable structure $(\mathcal{B}_1 \times \mathcal{B}_2, \tilde{\mathcal{B}}_1 \times \tilde{\mathcal{B}}_2)$ should be constructed as in **Definition 2.1**. Then, according to **Definition 2.2**, measurable random subsets (basic probability assignments) $\xi_1 : \Omega_1 \to \mathcal{B}_1 \times \Theta_2$, $\xi_2 : \Omega_2 \to \Theta_1 \times \mathcal{B}_2$ and $\xi_{12} : \Omega_1 \times \Omega_2 \to \mathcal{B}_1 \times \mathcal{B}_2$ can be defined. **Assumption 2.11** becomes: the combined probability field $(\Omega_1 \times \Omega_2, \mathcal{A}_1 \times \mathcal{A}_2, P_{12})$ has its combined net $\mathcal{D}_{12}^*$ and $\mathcal{A}_1 \times \mathcal{A}_2$ is sufficient to $\mathcal{B}_1 \times \mathcal{B}_2$ (and consequently, to $\mathcal{B}_1 \times \Theta_2$ and to $\Theta_1 \times \mathcal{B}_2$). **Lemma 3.1.2** also holds for

$$\xi_1(\omega_1) = G_{\xi_1}|_{\cdot \omega_1}, \quad \xi_2(\omega_2) = G_{\xi_2}|_{\cdot \omega_2}$$

and

$$\xi_{12}(\omega_1, \omega_2) = G_{\xi_{12}}|_{\cdot (\omega_1, \omega_2)},$$

where

$$G_{\xi_1}|_{\cdot \omega_1} \triangleq \{(\theta_1, \theta_2) | (\omega_1, \theta_1, \theta_2) \in G_{\xi_1}\},$$
$$G_{\xi_1} \in \mathcal{A}_1 \times (\mathcal{B}_1 \times \Theta_2);$$
$$G_{\xi_2}|_{\cdot \omega_2} \triangleq \{(\theta_1, \theta_2) | (\omega_2, \theta_1, \theta_2) \in G_{\xi_2}\},$$
$$G_{\xi_2} \in \mathcal{A}_2 \times (\Theta_1 \times \mathcal{B}_2)$$

and

$$G_{\xi_{12}}|_{\cdot (\omega_1, \omega_2)} \triangleq \{(\theta_1, \theta_2) | (\omega_1, \omega_2, \theta_1, \theta_2) \in G_{\xi_{12}}\},$$
$$G_{\xi_{12}} \in (\mathcal{A}_1 \times \mathcal{A}_2) \times (\mathcal{B}_1 \times \mathcal{B}_2),$$

respectively.

Suppose $\xi_1(\omega_1) \triangleq G^*_{\tilde{F}_1}|_{\cdot \omega_1} : \Omega_1 \to \mathcal{B}_1$ is the possibilistic evidence induced from a priori proposition $p_1 \triangleq X$ is $\tilde{F}_1$, $X \in \Theta_1$, where $G^*_{\tilde{F}}$ is the graph constructed as in **3.1**. We define the following operation:

**Definition 3.3.1** Let $G^e_{\xi_1} \triangleq \{(\omega_1, \theta_1, \theta_2) | (\omega_1, \theta_1) \in G^*_{\tilde{F}_1}, \theta_2 \in \Theta_2\}$, then $\xi^e_1(\omega_1) \triangleq G^e_{\xi_1}|_{\cdot \omega_1} : \Omega_1 \to \mathcal{B}_1 \times \mathcal{B}_2$ is called the *extending* of $\xi_1$ from $\Xi(\mathcal{A}_1, \tilde{\mathcal{B}}_1)$ to $\Xi(\mathcal{A}_1, \tilde{\mathcal{B}}_1 \times \tilde{\mathcal{B}}_2)$ (it is similar to the *refining* in (Shafer 1976)).

After extending, the conditional proposition is transfered into the product universe of discourse $\Theta_1 \times \Theta_2$, becoming $\vec{p}_{12}(X, Y) \triangleq$ If $(X, Y)$ is $(\tilde{F}_1, \Theta_2)$, then $(X, Y)$ is $(\Theta_1, \tilde{F}_2)$, $(X, Y) \in \Theta_1 \times \Theta_2$, which indicates the logical implication of the values taken by the two-dimensional language variable $(X, Y)$. In the two-dimensional universe of discourse $\Theta_1 \times \Theta_2$, we can also consider the two extreme cases, i.e., two-dimensional concepts $(\tilde{F}_1, \Theta_2)$ and $(\Theta_1, \tilde{F}_2)$ that are synonymous or antonymous in implication. For example, if the variables $X$ and $Y$ represent the *height* and *weight* of a person, respectively, then $\Theta_1 \triangleq [0cm, 300cm]$ and $\Theta_2 \triangleq [0kg, 200kg]$ are two different universes of discourse. The two dimensional concepts ("*Quite Tall*", $\Theta_2$) and ($\Theta_1$, "*Heavy*") may be considered as synonymous in two-dimensional implication, while ("*Very Short*", $\Theta_2$) and ($\Theta_1$, "*Heavy*") as antonymous in two-dimensional implication. Therefore, the possibilistic evidence $\xi^e_2 : \Omega_2 \to \mathcal{B}_1 \times \mathcal{B}_2$, which representes the posterior proposition $p_2 \triangleq Y$ is $\tilde{F}_2$, $Y \in \Theta_2$, should be induced from $\xi_2(\omega_2) \triangleq G^*_{\tilde{F}_2}|_{\cdot \omega_2} : \Omega_2 \to \mathcal{B}_2$ through extending, i.e., $\xi^e_2(\omega_2) \triangleq G^e_{\xi_2}|_{\cdot \omega_2}$,

where $\quad G^e_{\xi_2} \triangleq \{(\omega_2, \theta_1, \theta_2) | (\omega_2, \theta_2) \in G^*_{\tilde{F}_2}, \theta_1 \in \Theta_1\}$,

if $(\tilde{F}_1, \Theta_2)$ and $(\Theta_1, \tilde{F}_2)$ are synonymous in two-dimensional implication. If they are antonymous in two-dimensional implication, then $\xi^e_2 : \Omega_2 \to \mathcal{B}_1 \times \mathcal{B}_2$ should be induced from $\xi_2(\omega_2) \triangleq \overline{G^*_{\tilde{F}_2}}|_{\cdot \omega_2} : \Omega_2 \to \mathcal{B}_2$ through extending, i.e., $\xi^e_2(\omega_2) \triangleq \overline{G^e_{\xi_2}}|_{\cdot \omega_2}$,

where $\quad \overline{G^e_{\xi_2}} \triangleq (\Omega_2 \times \Theta_1 \times \Theta_2)/G^e_{\xi_2}$
$= (\Omega_2 \times \Theta_1 \times \Theta_2)/\{(\omega_2, \theta_1, \theta_2) | (\omega_2, \theta_2) \in G^*_{\tilde{F}_2}, \theta_1 \in \Theta_1\}$.

Morever, the result of a conditional inference is a measurable random subset from $(\Omega_1 \times \Omega_2, \mathcal{A}_1 \times \mathcal{A}_2, P_{12})$ to $(\mathcal{B}_1 \times \mathcal{B}_2, \tilde{\mathcal{B}}_1 \times \tilde{\mathcal{B}}_2)$. It is denoted as $\vec{\xi}_{12} \triangleq (\xi_1 \to \xi_2)$ whose property also depends on the selection of the combined probability measure $P_{12}$.

The conditional inference proposition $\vec{p}_{12}(X, Y) \triangleq$ If $(X, Y)$ is $(\tilde{F}_1, \Theta_2)$, then $(X, Y)$ is $(\Theta_1, \tilde{F}_2)$, $(X, Y) \in \Theta_1 \times \Theta_2$ and the union proposition $\vec{p}^*_{12}(X, Y) \triangleq (X, Y)$ is $not\ (\tilde{F}_1, \Theta_2)$ or is $(\Theta_1, \tilde{F}_2)$, $(X, Y) \in \Theta_1 \times \Theta_2$ are equivalent each other. To represent the two-dimensional fuzzy subset $not(\tilde{F}_1, \Theta_2)$, a body of possibilistic evidence $\overline{\xi^e_1}(\omega_1) \triangleq \overline{G^e_{\xi_1}}|_{\cdot \omega_1}$ will be induced, where $\overline{G^e_{\xi_1}}|_{\cdot \omega_1} \triangleq (\Omega_1 \times \Theta_1 \times \Theta_2)/G^e_{\xi_1}$. We have:

**Proposition 3.3.2**: $\vec{\xi}_{12} \triangleq (\xi_1 \to \xi_2)$ is a (measurable) random subset from $(\Omega_1 \times \Omega_2, \mathcal{A}_1 \times \mathcal{A}_2, P_{12})$ to $(\mathcal{B}_1 \times \mathcal{B}_2, \tilde{\mathcal{B}}_1 \times \tilde{\mathcal{B}}_2)$. If either $\tilde{F}_1$ or $\tilde{F}_2$ is not an empty set, $\vec{\xi}_{12}$ is a basic probability assignment.

We have the following relations in two-dimension similar to those in one dimension.

**Proposition 3.3.3**: If $(\tilde{F}_1, \Theta_2)$ and $(\Theta_1, \tilde{F}_2)$ are synonymous in two-dimensional implication, for all $(\theta_1, \theta_2) \in \Theta_1 \times \Theta_2$, there is

$$\mu^-_{\vec{\xi}_{12}}((\theta_1, \theta_2)) \leq \mu_{\vec{\xi}_{12}}((\theta_1, \theta_2)) \leq \mu^+_{\vec{\xi}_{12}}((\theta_1, \theta_2)).$$

If $(\tilde{F}_1, \Theta_2)$ and $(\Theta_1, \tilde{F}_2)$ are antonymous in two-dimensional implication, for all $(\theta_1, \theta_2) \in \Theta_1 \times \Theta_2$, the direction of the inequalities should be reversed. The center term in the above inequalities includes $\mu^I_{\vec{\xi}_{12}}((\theta_1, \theta_2))$ as a special case.

**Theorem 3.3.4**: If either $\tilde{F}_1$ or $\tilde{F}_2$ are normal fuzzy subset and $(\tilde{F}_1, \Theta_2)$ and $(\Theta_1, \tilde{F}_2)$ are synonymous in two-dimensional implication, then

$$Bel^-_{\vec{\xi}_{12}}(C) \geq Bel_{\vec{\xi}_{12}}(C) \geq Bel^+_{\vec{\xi}_{12}}(C);$$

$$Pl^-_{\vec{\xi}_{12}}(C) \leq Pl_{\vec{\xi}_{12}}(C) \leq Pl^+_{\vec{\xi}_{12}}(C)$$

for all $C \subset \Theta_1 \times \Theta_2$. If $(\tilde{F}_1, \Theta_2)$ and $(\Theta_1, \tilde{F}_2)$ are antonymous in two-dimensional implication, then the direction of the inequalities is reversed. The center term in the above inequalities includes $Bel^I_{\vec{\xi}_{12}}(C)$ and $Pl^I_{\vec{\xi}_{12}}(C)$ as a special case.

The operation $(\xi_1, \xi_2) \rightarrow \vec{\xi}_{12}$ is defined as *conditioning*. The result of this operation, $\vec{\xi}_{12}$, represents the conditional inference implication between the priori and posterior propositions $p_1 \triangleq X$ is $\tilde{F}_1$, $X \in \Theta_1$ and $p_2 \triangleq Y$ is $\tilde{F}_2$, , $Y \in \Theta_2$.

What are the implications of the different $P_{12}$'s? In the conditioning operation, there are two language variables $X$ and $Y$ taking their values under certain soft contraints provided by two fuzzy concepts $\tilde{F}_1$ and $\tilde{F}_2$. We can interpret that the $P_{12}$ represents the relation between the two variables $X$ and $Y$. In extreme cases, $X$ and $Y$ may be positively or negatively correlated, e.g., if $X$ represents a specific person's *weight* and $Y$ represents his *fatness*, then $X$ and $Y$ are positively correlative, $P_{12} = P^+_{12}$ is positive correspondent; if $X$ and $Y$ represent the *weight* and *thinness*, then they are negatively correlated, $P_{12} = P^-_{12}$ is negatively correspondent. Based on this understanding, the logical implication of **Theorem 3.3.4** is obvious.

**Note**: $X$ and $Y$ may be the same language variable, $\Theta_1$ and $\Theta_2$ may be the same universe of discourse, all the propositions and theorems in Subsection **3.3** hold for these special cases.

## 4 SUMMARY AND DISCUSSION

### 4.1 POSSIBILITY INFERENCE AND PROBABILITY INFERENCE

In Section **3**, we constructed a unified framework for the inference with possibilistic evidence. On this framework, the inference is carried out as probability inference. In Zadeh's possibility theory, possibility inference is carried out as fuzzy inference because possibility is interpreted as compatibility with (fuzzy) concepts. Many fuzzy inference operators were proposed for specific situations. On the framework developed in **3**, many fuzzy inference operators can be interpreted by taking different combined probability measures $P_{12}$. We have **Proposition 4.1**: For $P^+_{12}$, $P^I_{12}$ and $P^-_{12}$, if $\tilde{F}_1$ and $\tilde{F}_2$ are synonymous in implication, there are

1) $\mu^+_{\hat{\xi}_{12}}(\theta) = \max\{\mu_{\xi_1}(\theta), \mu_{\xi_2}(\theta)\} = \mu_{\xi_1}(\theta) \vee \mu_{\xi_2}(\theta);$

$\mu^+_{\hat{\xi}_{12}}(\theta) = \min\{\mu_{\xi_1}(\theta), \mu_{\xi_2}(\theta)\} = \mu_{\xi_1}(\theta) \wedge \mu_{\xi_2}(\theta).$

These are *Zadeh's union* and *intersection operators*;

2) $\mu^I_{\hat{\xi}_{12}}(\theta) = \mu_{\xi_1}(\theta) + \mu_{\xi_2}(\theta) - \mu_{\xi_1}(\theta)\mu_{\xi_2}(\theta);$

$\mu^I_{\hat{\xi}_{12}}(\theta) = \mu_{\xi_1}(\theta)\mu_{\xi_2}(\theta).$

These are the *probability sum* and *product operators* in fuzzy inference;

3) $\mu^-_{\hat{\xi}_{12}}(\theta) = \min\{\mu_{\xi_1}(\theta) + \mu_{\xi_2}(\theta), 1\};$

$\mu^-_{\hat{\xi}_{12}}(\theta) = \max\{\mu_{\xi_1}(\theta) + \mu_{\xi_2}(\theta) - 1, 0\}.$

These are the *bounded sum* and *bounded difference operators* in fuzzy inference.

4) $\mu^+_{\vec{\xi}_{12}}(\theta_1, \theta_2) = \min\{1 - \mu_{\xi_1}(\theta_1) + \mu_{\xi_2}(\theta_2), 1\}.$

This is *Lukasiewiez inference operator*.

5) $\mu^I_{\vec{\xi}_{12}}(\theta_1, \theta_2) = 1 - \mu_{\xi_1}(\theta_1) + \mu_{\xi_1}(\theta_1)\mu_{\xi_2}(\theta_2).$

This is the *probability inference operator*.

6) $\mu^-_{\vec{\xi}_{12}}(\theta_1, \theta_2) = \max\{1 - \mu_{\xi_1}(\theta_1), \mu_{\xi_2}(\theta_2)\}.$

This is *Zadeh's conditional inference operator*.

If $\tilde{F}_1$ and $\tilde{F}_2$ are antonymous in implication, then $\mu^+$ should be changed to $\mu^-$ (vise versa for $\mu^-$) in all the above equations.

From **Proposition 4.1**, we can conclude that the possibility inference and the probability inference can be unified on our framewok. The inference discussed in Section **3** maintains both the compatibility of concepts and the consistency of probability logic. Especially, $Bel^I_{\xi_1 \wedge \xi_2}(C)$ and $Pl^I_{\xi_1 \wedge \xi_2}(C)$ are the results by Dempster's Rule of Combination. In addition we have

**Corollary 4.2**: $\mu^+_{\bar{\xi} \vee \xi}(\theta) = 1$ and $\mu^+_{\bar{\xi} \wedge \xi}(\theta) = 0$ for all $\theta \in \Theta$. It means that $(\bar{\xi} \vee \xi)(\omega) = \Theta$ and $(\bar{\xi} \wedge \xi)(\omega) = \emptyset$, i.e. $\bar{\tilde{F}} \cup \tilde{F} = \Theta$ and $\bar{\tilde{F}} \cap \tilde{F} = \emptyset$, for $P^+_{12}$.

Because we consider $P^+_{12}$ as positive correlation between the two knowledge information sources in (union and intersection) combination inference, including the case of the two sources being the same one, **Corollary 4.2** agrees with intuition.

### 4.2 COMPATIBILITY WITH CONCEPTS AND IGNORANCE ON RANDOMNESS

The membership function $\mu_{\tilde{F}}(\theta) = \mu_{\xi}(\theta)$ and the commonality number $Q_{\xi}(C)$ may be considered as the compatibility of the values taken by the language variable $X$ with the concept $\tilde{F}$. In our investigation of inference, we have found that, in certain cases, the more compatible with concepts the values the language variable takes (i.e., the larger the membership function or the commonality number of the value taken by the variable), the less valuable the results obtained by inference, and vise versa. For example, the intersection of two concepts $\tilde{F}_1$ and $\tilde{F}_2$, which are synonymous in implication and are provided by two negatively correlated knowledge information sources, seems to be more valuable than that obtained from positively correlated sources, but we have $Q^-_{\tilde{\xi}_{12}}(C) \leq Q^+_{\tilde{\xi}_{12}}(C)$ for all $C \subset \Theta$ and $\mu^-_{\tilde{\xi}_{12}}(\theta) \leq \mu^+_{\tilde{\xi}_{12}}(\theta)$ for all $\theta \in \Theta$. Similar examples can be found in almost all the inequalities in Section 3. Could we conclude that the smaller the membership function (or commonality number) of a fuzzy set, the greater the information value it contains? Then, the empty set would contain the greatest information value. In the general sense, an emptyset does not contain any useful information. This seems to be a paradox. The problem is how to measure the randomness information value of inference results. It is reasonable that, the less the ignorence, the more valuable is the information on randomness. Therefore, the difference between *Plausibility* and *Belief*, $Pl(C) - Bel(C)$, may be an appropriate measure. This is suitable for union and conditioning operations and consistent with the concept of *inclusion for random sets* proposed by Dubois and Prade et al. (Dubois and Prade 1986, 1991; Yager 1986; Delgado and Moral 1987). However, it is not suitable for intersection operation. From a measure theory viewpoint, the belief function may be considered as an inner measure, and the plausibility function may be considered as an outer measure. In the case of the possibilistic evidence, these measures are induced by the constraints $\mu_\xi(\theta) = \mu_{\tilde{F}}(\theta), \theta \in \Theta$. This view provides an interesting connection with the measurability of subsets. Some researchers have paid attention to this topic, e.g. (Fagin and Halpern 1991). We believe that the measurability of a subset (proposition) is closely related to the grain size or resolution of the properties specifying the subsets.

### 4.3 COMPUTATIONAL COMPLEXITY OF INFERENCE

It is difficult to determine the correlation between the knowledge information sources. It is also difficult to determine whether concepts are synonymous or antonymous in their implications. In spite of these, the inequalities obtained in Section 3 provide the bounds of inference results (**Note**: Even when two concepts cannot be determined to be synonymous or antonymous in implication, inequalities similar to those in Section 3 can be obtained with a little revision, i.e., the bounds on inference results still hold). Probability inference, especially, the combination inference for evidence (e.g., using *Dempster's Rule of Combination*), usually has a high computational complexity. To carry out the combination in an infinite universe of discourse, the amount of computation needed tends to grow exponentially with the accuracy required. It would be impossible to complete such logical inference of combination in practice. However, determination of $P_{12}^+$ and $P_{12}^-$ is much easier and the complexity of computation is much lower if $P_{12}^+$ and $P_{12}^-$ are used to combine $\xi_1$ and $\xi_2$. We believe that $P_{12}^+$ and $P_{12}^-$ may give the lower and upper bounds (or vise versa) of the combination results of the other $P_{12}$'s. It may probably provide a new technique to the development of fast inference mechanism with possibilistic evidence.